\documentclass[runningheads]{llncs}
\newcommand{\tabincell}[2]{\begin{tabular}{@{}#1@{}}#2\end{tabular}} 
\usepackage{graphicx}
\usepackage{comment}
\usepackage{amsmath,amssymb} 
\usepackage{stfloats}
\usepackage{float}
\usepackage{wrapfig}
\usepackage{subfigure}
\usepackage{color}
\usepackage{booktabs} 
\usepackage{multirow}
\usepackage{color}

\makeatletter 
\newcommand\figcaption{\def\@captype{figure}\caption} 
\newcommand\tabcaption{\def\@captype{table}\caption} 
\makeatother

\definecolor{todo}{rgb}{1,0,0}  

\begin{document}
\pagestyle{headings}
\mainmatter
\def\ECCVSubNumber{3878}  

\title{Music2Dance: DanceNet for Music-driven Dance Generation} 

\titlerunning{Music2Dance} 
\authorrunning{W. Zhuang et al.} 
\author{Wenlin Zhuang\inst{1} \and
Congyi Wang\inst{2} \and
Siyu Xia\inst{1} \and
Jinxiang Chai\inst{2,3} \and
Yangang Wang\inst{1}
}
\institute{Southeast University
\email{wlzhuang@seu.edu.cn, xia081@gmail.com,yangangwang@seu.edu.cn}
\and
Xmov
\email{artwang007@gmail.com}\\
 \and
Texas A\&M University\\
\email{jchai@cs.tamu.edu}}

\maketitle

\begin{abstract}
Synthesize human motions from music, i.e., music to dance, is appealing and attracts lots of research interests in recent years. It is challenging due to not only the requirement of realistic and complex human motions for dance, but more importantly, the synthesized motions should be consistent with the style, rhythm and melody of the music.
In this paper, we propose a novel autoregressive generative model, DanceNet, to take the style, rhythm and melody of music as the control signals to generate 3D dance motions with high realism and diversity. To boost the performance of our proposed model, we capture several synchronized music-dance pairs by professional dancers, and build a high-quality music-dance pair dataset. Experiments have demonstrated that the proposed method can achieve the state-of-the-art results. 

\end{abstract}

\section{Introduction}\label{sec:introduction}

As an art of human motion, dance plays an important role in culture, 
sports and related fields, e.g., art programs, rhythmic gymnastics, 
figure skating. Conventionally, 
dance is always involved with music to enhance artistic appeal, 
and the combination of music and choreography needs careful design and meticulous arrangement. 
In general, music and choreography should not only show the artistic quality of dance, but also need to reflect the content of music (music-consistency). 
Particularly, the artistic quality requires the dance to be realistic and diverse, and the music-consistency requires style-consistence and melody-matching between human motions and music.
Performing efficient and fully automatic choreography with music is always 
challenging, and thus becomes a hot research topic 
in the filed of computer vision and computer graphics~\cite{cardle2002music,tang2018dance,lee2019dancing,yan2019convolutional}. 
In this paper, we focus on the problem of automatic choreographing with music.

Historically, motion generation and synthesis is often solved by data-driven approaches\cite{kovar2008motion,min2012motion,fragkiadaki2015recurrent}. 
However, to synthesize complex human motions coinciding with music, two main challenges need to be addressed, i.e., high quality motion data as well as appropriate motion models. 
Early researchers\cite{kovar2008motion,min2012motion} took
the idea of stitching motion to generate long-term motions based on a large amount of data. 
With the development of deep learning\cite{lecun2015deep,krizhevsky2012imagenet},
Recurrent Neural Network(RNN)\cite{fragkiadaki2015recurrent,martinez2017human,gopalakrishnan2019neural} has been proposed in recent years to model human motions. However, RNN based methods can easily fall into a static pose due to error accumulation(freeze)\cite{li2017auto}, especially when the input is not identical to the training data or noises are presented. 
In order to solve the problem, temporal convolution\cite{ginosar2019learning} was proposed to generate simple gesture, which is robust to noise, but it can only model simple gestures. 
It is noted that most of the existing methods can only model simple and regular locomotion, and it is difficult to model complex and diverse dance motions, which can not be used to synthesize human motions from controllable music.  
Recently, Lee et al.\cite{lee2019dancing} used VAE and GAN to model 2D dance motion, and Tang et al.\cite{tang2018dance} trained an LSTM-autoencoder to generate 3D dance motion directly from music features, but their generated dance motion is far from realistic and diverse. 

Aiming at the technical challenges to model complex human motions for music, we propose a novel autoregressive model, DanceNet. Different from previous methods\cite{tang2018dance}\cite{lee2019dancing} which  directly map music features to human motions, we take the strategy to perform music features as control signal. Specifically, to improve the model adaptation for complex human motions synthesis, our key idea is to introduce dilated convolution as well as gated activation to  build a compact model. To ease the network training and adaptation, we adopt a special loss function, Gaussian Mixture Model(GMM) loss, which is different from widely used loss function(MSE loss) in many motion generation models\cite{fragkiadaki2015recurrent,li2017auto,martinez2017human,holden2016deep}.
Our strategies can generate complex and diverse human motion and improve the network efficence for human motion generation without largely increasing the network parameters.
Compared with RNN model\cite{fragkiadaki2015recurrent,martinez2017human,lee2018interactive}, the convolution-based model we proposed is more robust. For general temporal convolution\cite{ginosar2019learning}, it is very difficult to model complex and diverse dance due to its small receptive field and low model complexity. 
Naively stacking many temporal convolution layers to increase the receptive field can make the model difficult to train. Therefore, we adopt dilated convolution to increase the receptive field without changing the depth of the network. Inspired by WaveNet\cite{oord2016wavenet} and PixelCNN\cite{van2016conditional}, we add the gated activation unit to increase the complexity and improve the modeling ability for complex, irregular and diverse dance motion. 
In order to improve the robustness of the model and the diversity of the generated dance motion, we adopt GMM loss to train our DanceNet and add gaussian noise to groundtruth during the training phase. During the testing phase we can sample from the model output, which increases the diversity of the generated dance. In our motion data representation, we consider the 
error accumulation from root to end-effector and the foot sliding, so we add end-effector position and foot contact to motion representation(our model is contact-aware). For the control signals, we take the musical style, rhythm and melody as control signals. The musical style determines the dance type, and the rhythm and melody determine the dance rhythm and other characteristics, e.g., dance amplitude, velocity. It is completely different from the existing methods\cite{lee2018listen,tang2018dance}, they extract the spectrum and other audio features as input. Our music features are more closely related to the dance motion.


Beyond the above technical challenges, we captured several high-quality music-dance motion pair dataset to boost the performance of music dance generation. Existing 3D human motion datasets\cite{CMUdata,SFUdata,Mixamo,yun2012two} are all about simple locomotion, lacking dance motion, and more importantly, they are not the music-dance pair datasets. The datasets which contain dance motions also have many drawbacks.  Lee et al.\cite{lee2018listen} only constructed 2D dance motions. Tang et al.\cite{tang2018dance} collected about $1.5$ hours of dance motion by motion capture devices, but the dance motions are unrealistic, simple(not diverse), and do not match the music. The lack of high-quality music-dance pair data makes it difficult for the topic to take a step. To solve this problem, we build a high-quality music-dance pair dataset. Our dataset contains two typical dance types, modern dance about 26.15min(94155 frames, FPS=60), and curtilage dance about 31.72min(114192 frames, FPS=60), 
both are accurately aligned with the corresponding music. The 3D dance motion we collect contains finger motion, which can make the dance more realistic and diverse. The dataset will be public available in the future.

To demonstrate the effectiveness of our DanceNet, we compared it against different sequence models: LSTM-autoencoder\cite{tang2018dance}, 
Temporal Convolution\cite{ginosar2019learning}, and SOTA LSTM(trained with our GMM loss)\cite{lee2018interactive}. The results show that our mehod can achieve SOTA result, and the dance motions generated by our method are not only realistic and diverse, but also are music-consistent. Our results are shown at~\url{https://youtu.be/bTHSrfEHcG8}.


The main contributions of this work include:
\begin{itemize}
    \item We propose a novel autoregressive generative model, DanceNet, takes the musical style, rhythm and melody as control signals, and combines with GMM loss. Our method can generate realistic, diverse, and music-consistent dance motion.  
    \item We build a high-quality music-dance pair dataset, and the dance motion includes finger motion.
    \item Compared with other sequence models, the experimental results show that our method can achieve SOTA result.
\end{itemize}

\section{Background}\label{sec:background}

\textbf{Music signal representation}. 
In \cite{lee2018listen,tang2018dance,yalta2018weakly}, 
the researchers used Mel spectrum, Mel Frequency Ceptral Coefficient(MFCC) or
short-time Fourier transform (STFT) spectrum as a music feature to represent music.
Although it is widely used in speech recognition, it is not applicable to music. 
Because it is a low-level feature that all audio contains, 
it is not suitable as the music feature. 
The most basic features in music are beat, rhythm and melody. 
More critically, this is the most important dependency in dance generation. 
In the music information retrieval, most of the work is about how 
to extract the music features. 
Onset can expresses the beginning of music notes and it is the most basic expression form of music rhythm\cite{hawthorne2017onsets,eyben2010universal,bock2012online}.
Beat is another form of rhythm, 
and there is a lot of work on beat 
detection\cite{bock2016joint,krebs2016downbeat,krebs2013rhythmic}. 
Melody, one of the most important music features, 
can be expressed via 
chroma feature\cite{gomez2006tonal,korzeniowski2016feature,muller2011chroma}. 
Most importantly, chroma feature is highly adaptable to changes in timbre and instruments. 
Therefore, we adopt onset, beat and chroma as the music features to represent music.


\vspace{+2mm}\textbf{Human motion generation}. The early methods on motion generation can be divided into 
Hidden Markov Models(HMMs) \cite{bowden2000learning,brand2000style}, 
statistical dynamic model \cite{li2002motion,pavlovic2001learning,chai2007constraint,lau2009modeling,xia2015realtime} 
and low-dimensional statistical model\cite{chai2005performance} 
for human poses. The most famous approach is motion graph\cite{kovar2008motion,min2012motion,safonova2007construction}.The above methods use the idea of stitching to generate motion, which requires a large dataset, and it can only be used for simple regular locomotion.
Recent work focus on deep learning methods to model human motion based on RNN\cite{fragkiadaki2015recurrent,jain2016structural,li2018convolutional,martinez2017human,gopalakrishnan2019neural}. However, most of their methods are difficult to generate long-term motion due to error accumulation. The methods in \cite{holden2016deep,lee2018interactive} can generate long-term simple locomotion, but cannot model complex and diverse dance motion. Li et al.\cite{li2017auto} proposed auto-conditioned LSTM to solve error accumulation and generate long-term dance motion, but the generated motion is unrealistic, simple(not diverse), and the model cannot be controlled with music. 
The method we proposed can solve this problem and generate realistic, diverse and music-consistent dance motion.


\textbf{Autoregressive model.} In motion generation, RNN(LSTM)\cite{lee2018interactive,fragkiadaki2015recurrent,li2017auto} is usually used as an autoregressive model to model motion. But LSTM is not robust to noise. We propose an autoregressive model based on dilated convolution, and inspired by PixelCNN\cite{van2016conditional} and WaveNet\cite{oord2016wavenet}, we add the gated activation unit to our model. Our experiments demonstrate that the proposed model is more robust to existing RNN-based method\cite{lee2018interactive} and can produce more realistic and diverse dance motion.



\section{Dataset}\label{sec:dataset}

\begin{table}[t] 
    \begin{minipage}{0.57\textwidth} 
    \centering
    \caption{\textbf{Comparison with existing 3D motion dataset} in the following aspects: whether it contains a lot of dance motion(Y:yes,N:no), dance artistry(H:high, M:midddle, L:low), music-consistency(H:high, M:midddle, L:low, N:no music), and whether it contains finger motion(Y:yes,N:no).}
    \begin{tabular}{l|c|c|c|c} 
    \hline
      Dataset & Dance  & Artistry  & \tabincell{c}{Music- \\ Consistency} & Finger\\
	  \hline
      CMU\cite{CMUdata}     & N & - & - & N \\
      SFU\cite{SFUdata}     & N & - & - & N \\
      Mixamo\cite{Mixamo}  & N & - & - & N \\
      Tang \cite{tang2018dance}  & Y & L & M & N \\
      \hline
      Ours  & \textbf{Y} & \textbf{H} & \textbf{H} & \textbf{Y} \\

       \hline 
    \end{tabular}  
    \label{table:data_compare} 
  \end{minipage}
  \begin{minipage}{0.4\textwidth} 
  \centering 
    \includegraphics[width=5cm]{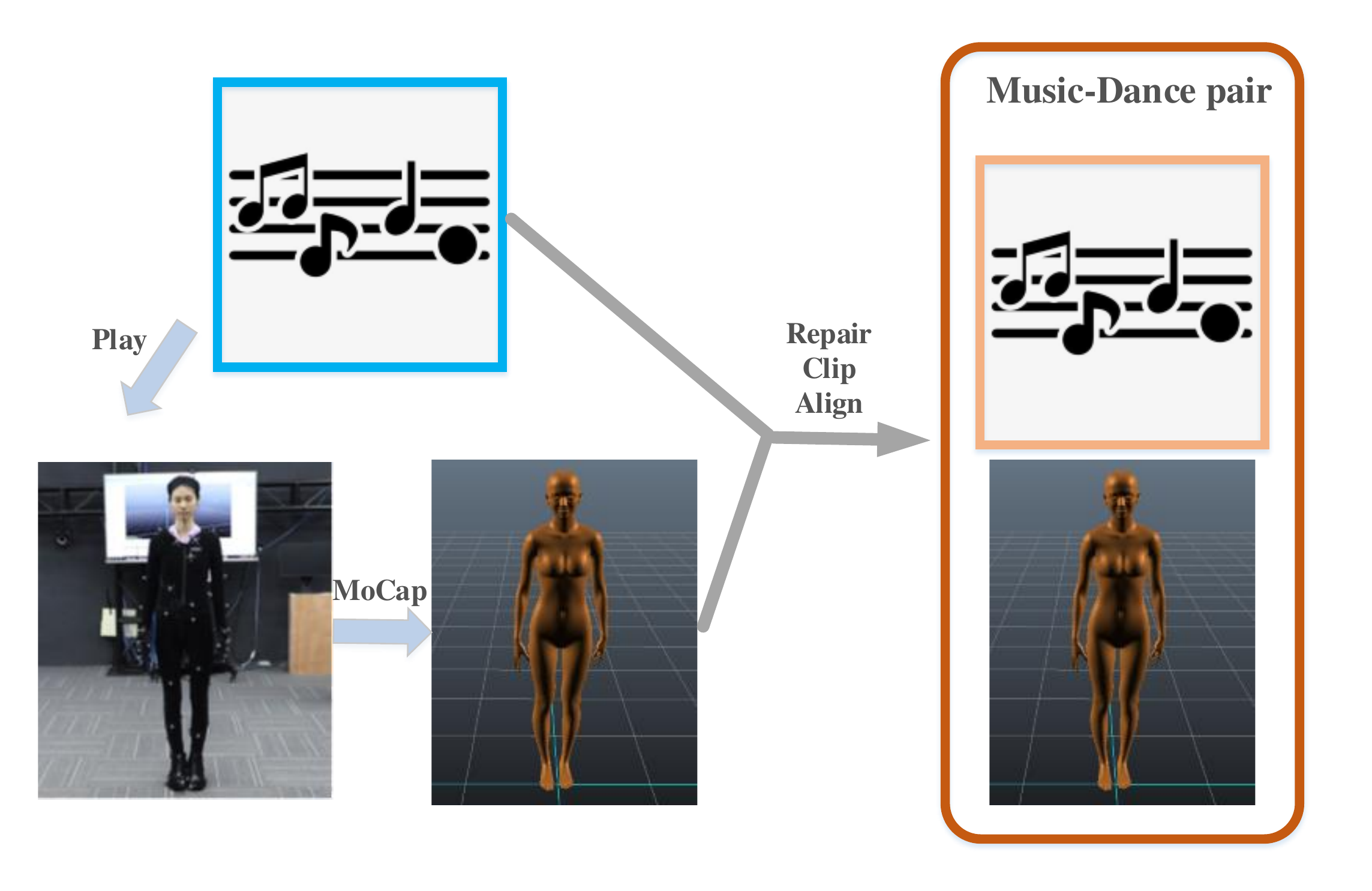} 
    \figcaption{\textbf{Our music-dance pair data collection process}. There are three steps: 
     	1) playing music and professional dancers dancing with music, 
     	2) motion capture and collection system collecting motion,
     	3) repairing motion, clipping and registration according to music.}
    \label{fig:collection_data} 
  \end{minipage}%
\end{table}

    


\textbf{Music data.}
In order to generate style-consistent dance from music, 
we need to collect a lot of music data to train a classification model to get musical style.
The musical style contains two types: smoothing music and fast-rhythm music.  
We downloaded the music songs from the music website, 
and then carefully distinguished the musical style. 
Finally, we collected about 35.7 hours music songs. 
18.2 hours are smoothing music (suitable for modern dance), 
and the others are fast-rhythm music (suitable for curtilage dance).


\textbf{Music-Dance pair.}
To the best of our knowledge, there are few datasets of dance motions. 
CMU motion-capture \cite{CMUdata}, SFU motion-capture \cite{SFUdata}, and Mixamo \cite{Mixamo} are most about walking, running and jumping. They are simple motions, lacking dance motion, and more importantly, there is no corresponding music.
Tang et al.\cite{tang2018dance} attempted to collect dance motions with motion capture devices, but the quality of dance motions is low, and their captured motions do not match the corresponding music (not music-consistent). In order to achieve the artistic(realistic, diverse), music-consistent and high-quality dance motion generation, we have collected high-quality music-dance pair data. 
The comparison with existing 3D motion dataset is shown in Table ~\ref{table:data_compare}.
Our data collection process includes three steps: 
1) playing music and letting the professional dancers to dance with music, 
2) collecting human motions with a motion capture system,
3) repairing, clipping and registering motions according to music, 
and finally producing music-dance pair data, as shown in Figure ~\ref{fig:collection_data}.


We asked two professional dancers (a man and a woman) to 
collect the modern and curtilage dance motions respectively.
To get high-quality dance motions, we spent a lot of time on data repair and alignment,
and then we obtained modern dance motion about 26.15min (94155 frames, FPS=60) 
and curtilage dance motion about 31.72min (114192 frames, FPS=60), for a total of 208,347 frames, 57.87min. 
Notably, each frame of our dance motion contains 55 joints including fingers, 
which can help to improve the artistry of the dance motion.

\section{Methodology}\label{sec:Methodology}



\begin{figure}[t]
	\centering
	\includegraphics[width=\linewidth]{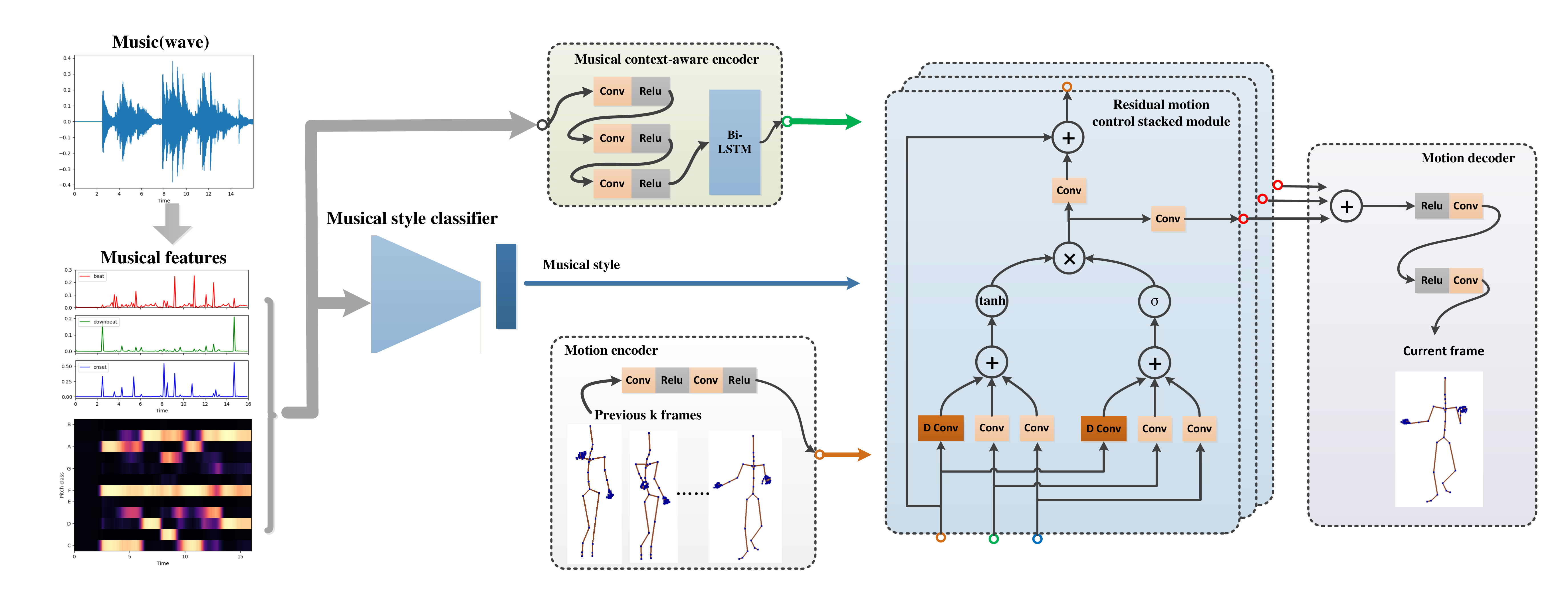}
	\caption{\textbf{Our framework}. First we extract the musical rhythm and melody (music features) and classify the musical style by the musical style classifier. DanceNet takes the musical style, rhythm and melody as control signals to generate dance motion. 
	DanceNet consists of four parts: musical context-aware encoder, motion encoder, residual motion control stacked module, motion decoder. The "Conv" is 1D convolution, and the "D Conv" is dilated convolution.}
	\label{fig:framework}
\end{figure}

\textbf{Framework.}
Our goal is to generate realistic, diverse and music-consistent dance motion. 
In order to achieve the goal, 
we propose a novel autoregressive generative model, DanceNet, takes the musical style, rhythm and melody as control signals, and trained by GMM loss. Our framework is shown in Fig~\ref{fig:framework}.


\subsection{Music feature and music-consistency analysis}\label{sec:music_feature_extraction}

\begin{figure}[t]
    \centering
    \subfigure[music feature]
    {\includegraphics[width=5.7cm, angle=0]{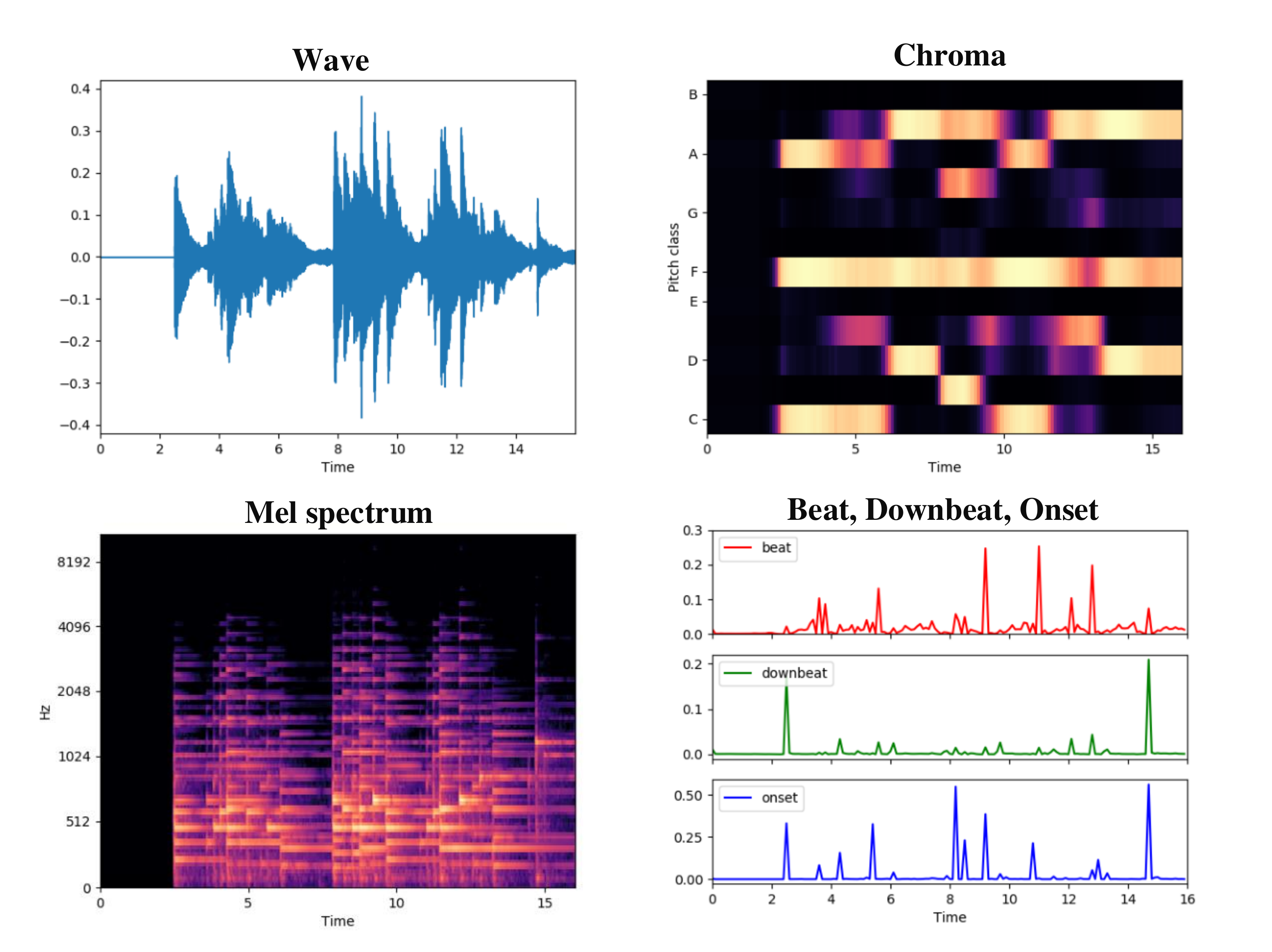}}
    \subfigure[rhythm-consistency analysis]
    {\includegraphics[width=6cm, angle=0]{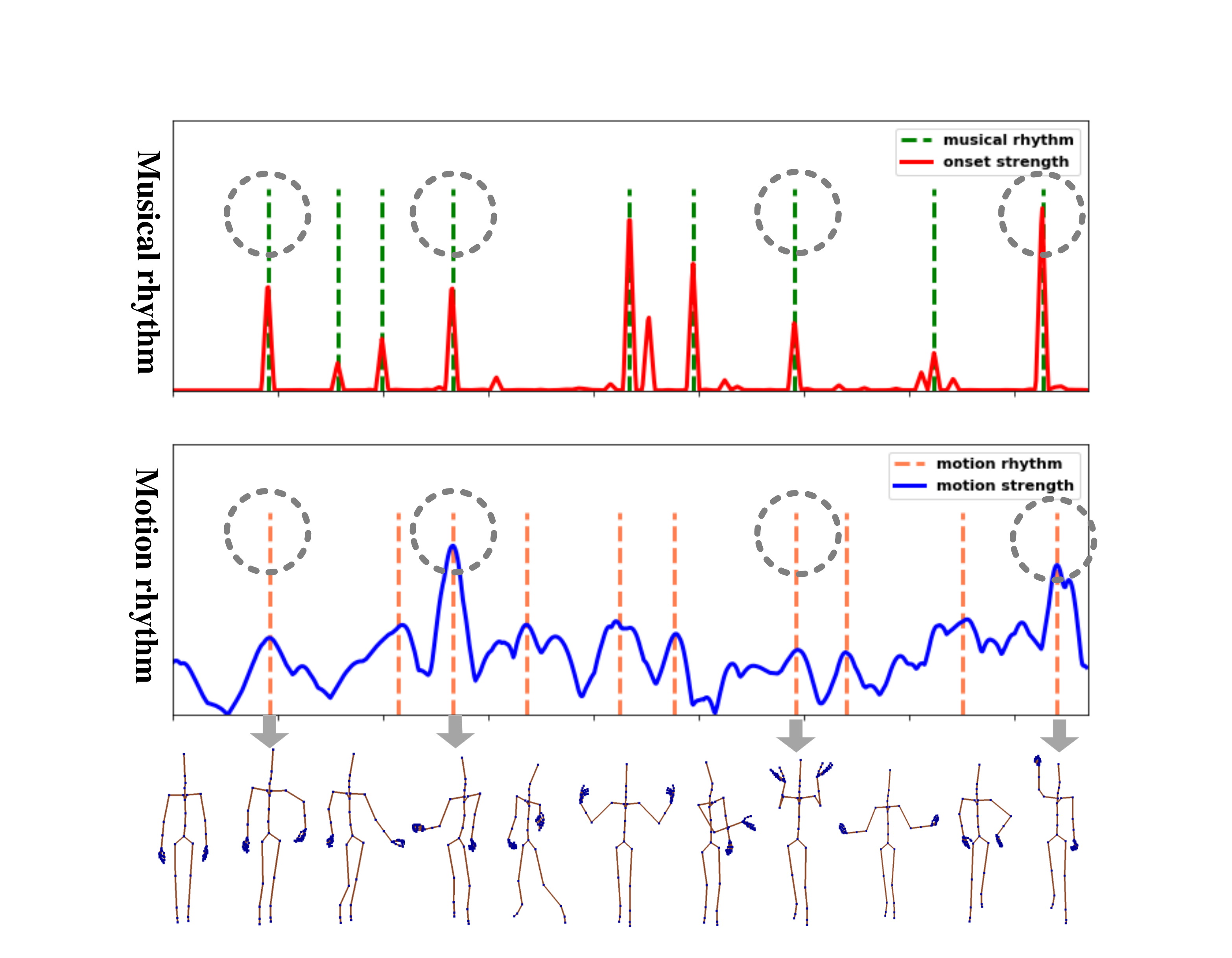}}
    \caption{\textbf{Different music feature and rhythm-consistency analysis}. (a)Music is represented by wave, with Mel spectrum as its basic feature. Chroma, beat(beat, downbeat) and onset are its high-level features. (b) We extract the musical rhythm from onset, and then extract dance rhythm from the corresponding dance motion. The dance rhythm matches the musical rhythm basically. }
    \label{fig:music_feature_analysis}
\end{figure}

Instead of using Mel spectrum or other acoustic features, 
we adopt high level music features: rhythm and melody. We adopt onset and beat to represent musical rhythm, and chroma to represent musical melody. We rely on madmom \cite{bock2016madmom} to extract the onset, beat, chroma. 
The onset feature is a 1D vector, and the value represents 
the onset strength.
The beat feature, including the beat and downbeat, 
 can express the rhythm information. 
 The chroma feature closely relates to the twelve different pitch classes 
 and can characterize the musical melody, as shown in Figure ~\ref{fig:music_feature_analysis}(a). 
 The frame per-second (FPS) of the three music features is 10, 
 and the dimension of each frame is 15 (onset 1, beat 2, chroma 12). 
 
 As we explained above, our goal is to generate a music-consistent dance motion, which is reflected in  matching with musical style, rhythm and melody.
 Style-consistency is basically determined by the performance of the musical style classification. Intuitively, the dance rhythm has to match the musical rhythm. In order to analyze the consistency between them, we extracted the musical rhythm and dance rhythm in our music-dance pair dataset, respectively. The musical rhythm is directly extracted from the onset, and the dance rhythm is extracted from the motion strength (the sum of the velocity of the joints), similar to \cite{lee2019dancing}, as shown in Fig.~\ref{fig:music_feature_analysis}(b). The musical rhythm and dance rhythm basically match, which shows that they are indeed strongly related. In addition, it can also be used as an evaluation criterion to evaluate our results.


\subsection{Musical style classification}\label{sec:music_attribute}
The musical styles include two types: the music that is suitable 
for modern dance (e.g., smoothing music) 
and the music that is suitable for curtilage dance(e.g., fast-rhythm music). 
In our method, the input of Musical style classifier $F_{m}$ is 
the music features $x_{m}$ (onset, beat, chroma mentioned in 
Section ~\ref{sec:music_feature_extraction}), and the output is the musical style $m_{s}$ (two categories, represented by a one-hot code). 


The input $x_{m}$ is computed within a sliding window, and the window size must be carefully considered, 
because the music attribute is not determined by a short music clip (2-3 seconds). 
In our experiment, the window size is 30 seconds.
Inspired by convolution RNN\cite{choi2017convolutional}\cite{hawthorne2017onsets} and temporal convolution\cite{pavllo20193d}, we adopt 3 temporal convolution layers 
and 1 Bi-LSTM (Bi-directional LSTM) layer as a high-dimensional feature extractor, 
and finally it is classified 
by a fully connected layer.
The kernel size of temporal convolution is set to 2, 
and the feature channels are in order: 1-16-24-32. 
We adopt cross entropy as loss function.


\subsection{Motion representation}\label{sec:motion_representation}
Each frame in the motion data contains 55 joints: 
one is for root joint, whose motion is represented 
by translation and rotation related to the world coordinate 
($t_{x}, t_{y}, t_{z}, r_{x}, r_{y}, r_{z}$), 
and the remaining joints
are represented by the rotation related to 
their parent joints ($r_{jx}, r_{jy}, r_{jz}$, $j$ is the joint index). 
To better describe the motion feature, 
we modify the root joint representation. 
We use the relative rotation $\Delta r_{y}$ 
between current frame and previous frame for the rotation around 
Y-axis(vertical axis of human pose), 
and the $x$, $z$ translation of the root joint are defined on the local coordinate of 
previous frame ($\Delta t_{x}, \Delta t_{z}$), similar to \cite{holden2016deep}.
There is a great advantage: no matter where the last frame moves to 
and which direction it faces, 
our method can describe the next frame motion, 
which indicates the invariant of our data representation. 
The joint rotation motion thus can be described as follows:
\begin{equation}
x_{rot}=({[\Delta t_{x}, t_{y}, \Delta t_{z}, \Delta r_{y}, r_{x}, r_{z},}\\ 
{r_{2x}, r_{2y}, r_{2z}, ..., r_{jx}, r_{jy}, r_{jz}]})\\
\label{equ:motion_rot}
\end{equation}

However, if we adopt such a representation, 
there would generate large accumulation errors. 
Because the joints from the root to the end-effectors are rotated relative to the parent joint, a large error appears on the end-effector position 
if the rotations are inaccurate, 
which greatly reduce the quality of the dance motion. 
To solve this problem, 
we add the end-effector position into the motion representation, 
so that our model can predict the end-effector position in each predicted frame. 
This effectively helps to eliminate the accumulation error. 
The end-effector motion is described as:

\begin{equation}
p_{end}=[p_{1x}, p_{1y}, p_{1z}, ..., p_{kx}, p_{ky}, p_{kz}]
  \label{equ:motion_endp}
\end{equation}
where $k$ is the end-effector index. 
In our method, we use left/right $toe\_end$ as foot end-effectors, $head\_end$ as head end-effector.
Since our motion data includes finger motion and there are too many end-effectors in the hand,
 we use left/right hand as the end-effectors of two hands, respectively. 
In addition, considering foot sliding, we add foot constraints to the motion feature, which makes our DanceNet contact-ware, as in \cite{xia2015realtime,holden2016deep}.
  We adopt a 2d vector $c_{foot}$ to describe 
  whether the left/right foot are in contact and fixed to the ground. 
  By detecting the left/right $toe\_end$ position and speed of each frame, 
  the ground-truth of the foot constraints are obtained. 
  Finally, the motion feature $x$ includes:$x=[x_{rot}, p_{end},c_{foot}]$.

\subsection{DanceNet structure}\label{sec:dance_generation}

Unlike LSTM\cite{lee2018interactive,fragkiadaki2015recurrent,li2017auto}, our DanceNet can model the conditional distribution $p({x}| x_{m}, m_{s})$ to improve model robustness. It takes the music features(musical rhythm and melody) $h_{m}$ and the musical style $m_{s}$ (dance type) as control signals to output the conditional distribution $p({x}| x_{m}, m_{s})$. Therefore, our DanceNet can be described as:
\begin{equation}
	p({x}| x_{m}, m_{s})=\prod_{t=1}^{T}p({x}_{t}|{x}_{t-k-1},...,{x}_{t-1}, x_{m,t}, m_{s})
	  \label{equ:dancenet_model_train}
\end{equation}
where $k$ is the receptive field. Specifically, the structure of our DanceNet is divided into the following parts: musical context-aware encoder, motion encoder, residual motion control stacked module, and motion decoder, as shown in Fig~\ref{fig:framework}.

\textbf{Musical context-aware encoder.}
Since the motion and music are two different modalities, directly taking music features as control signals would cause difficulty in feature fusion and reduce modeling capabilities.
We combine temporal convolution 
and Bi-LSTM as the music encoder. It guarantees that the music-context-code can completely represent the musical rhythm and melody at every moment, while taking into account the musical context information. In addition, since dance motion is smooth, the Bi-LSTM is added after time convolution to fuse the context information and ensure that the music-context-code is smooth(there is jitter in the musical rhythm and melody, as shown in  Fig~\ref{fig:music_feature_analysis}(a)).
Similar to Sec~\ref{sec:music_attribute}, the music features $x_{m}$ input to music encoder via the sliding window. In order to improve the training speed, the window size is fixed during the training phase, and we set it to 8s. In the inference phase, the music-context-code of each clip is extracted via the sliding window (overlapping 2s) and then stitched to form a complete music-context-code.

\textbf{Motion encoder.} We stack two "Conv1D+Relu" module as motion encoder to encode the past $k$ frames. The convolution kernel is set to 1, which ensures that each frame motion code(512 channels) is independent.

\textbf{Residual motion control stacked module}. To increase the receptive field, we adopt dilated convolution in our module. Inspired by the gate activation unit in PixelCNN\cite{van2016conditional} and WaveNet\cite{oord2016wavenet}, we add the gated activation unit and combine it with the dilated convolution to form the module.The gated activation helps to improve the interaction between different features and model complexity. We adopt two dilated convolution to compute filter and gate features, respectively. The structure increases the capacity of our module by decoupling the features, and it is better than sharing the same dilated convolution for modeling dance motion.
In addition, we use Conv1D(kernel size is 1) for all control signals. The features from the dance motion, musical style and music-context-code are fused by
adding operations. It should be noted that each time a dilated convolution is used, padding zeros is required to ensure that it only depends on the previous frames. In our experiments, we stacked 20 residual motion control modules with 32 feature channels and maximum dilation coefficient of 5.

\textbf{Motion decoder}. We build the decoder by two stacked "Relu+Conv1D" to map the fused features from residual motion control stacked module to the predicted the distribution of dance motion.

\subsection{GMM loss}\label{sec:gmm_loss}
In order to predict the distribution of current frame, 
Gaussian Mixture Model (GMM) is adopted to model the probilistic distribution of the motion feature in the current frame.
The distribution of GMM is:
\begin{equation}
	Pr(x_{t})=\sum_{i=1}^{N}\omega _{i}\mathbb{N}(x_{t} | \mu _{i},\Sigma _{i})
	  \label{equ:gmm_dis}
\end{equation}
GMM model requires $\sum_{i=1}^{N}\omega _{i}=1,0\leq \omega _{i}\leq 1$, $\Sigma _{i}>0$.
$N$ is the number of gaussian mode, $\mu _{i}, \Sigma _{i}$ is the mean vector and covariance matrix, respectively. To meet the requirements, we define the output of our model as:
$\hat{\omega}_{i},\hat{\mu }_{i,j},\hat{\sigma }_{i,j}$ ($\hat{\omega}_{i}\subseteq \hat{\omega},
\hat{\mu }_{i,j}\subseteq \hat{\mu },\hat{\sigma }_{i,j}\subseteq \hat{\sigma }$,
$j$ is the dimension index of motion feature ($x_{rot}, p_{end}$), the model output includes: 
$\hat{\omega}$, $\hat{\mu }$, $\hat{\sigma }$, $\hat{c}_{foot}$), 
and the requirements can be satisfied:
\begin{equation}
	\omega _{i}=\frac{e^{\hat{\omega}_{i}}}{\sum e^{\hat{\omega}_{i}}},{\mu }_{i,j}=\hat{\mu }_{i,j},\Sigma _{i,j}=e^{\hat{\sigma }_{i,j}}
	  \label{equ:gmm_dis_para}
\end{equation}
The loss function is defined as the negative log likelihood:
\begin{equation}
L_{gmm}=-logPr(x_{t} | \omega _{i},{\mu }_{i},\Sigma _{i}) 
=-log\sum_{i=1}^{N}\omega _{i}\mathbb{N}(x_{t} | \mu _{i},\Sigma _{i})
\label{equ:gmm_loss}
\end{equation}
In our experiment, $N$ is set to 1. The negative log GMM loss calculates the joint rotation motion $x_{rot}$ and the end-effector position motion $p_{end}$. 
For foot constraints $c_{foot}$, 
binary cross entropy loss ($BCE$ loss) is adopted:
\begin{equation}
	L_{foot}=BCE(\hat{c}_{foot},c_{foot})
	  \label{equ:foot_loss}
\end{equation}
so the loss fuction $L_{gen}$ in training phase is,
\begin{equation}
	L_{gen}=L_{gmm}+ \lambda * L_{foot}
	  \label{equ:foot_loss}
\end{equation}
To balance the two loss functions, we set a parameter $\lambda$ and set it to 0.1 in our experiment.
In the inference phase, the predicted motion feature $\hat{x}_{rot}, \hat{p}_{end}$ 
can be sampled from the GMM model, and the predicted motion feature can be described as follows,
\begin{equation}
[\hat{x}_{rot}, \hat{p}_{end}]= S_{gmm}(\hat{\omega},\hat{\mu },\hat{\sigma })
\label{equ:gmm_sample}
\end{equation}
\begin{equation}
	\hat{x}=[\hat{x}_{rot}, \hat{p}_{end}, \hat{c}_{foot}]
	\label{equ:output_motion}
	\end{equation}
After obtaining the predicted dance motion, we perform post-processing, and the details are described in the supplementary materials.

\section{Experiment}\label{sec:Experiment}
The implementation details are described in supplementary materials. 
In this section, we demonstrate our 
approach by evaluating our results (Section ~\ref{sec:our_result}) 
and analyzing our methods (Section ~\ref{sec:discuss}).

\setlength{\tabcolsep}{4pt}
\begin{table}[t]
\begin{center}
\caption{
	Musical style classification comparision on testing data (Accuracy).
}
\label{table:music_attribute}
\begin{tabular}{l|c}
\hline
Method & Accuracy (\%)  \\
\hline
Convolution RNN\cite{choi2017convolutional} & 89.6 \\
Baseline (FC + Bi-LSTM) & 88.0   \\
\textbf{Ours (Temproal Conv + Bi-LSTM)} & \textbf{92.1} \\
\hline	
\end{tabular}
\end{center}
\end{table}
\setlength{\tabcolsep}{1.4pt}

\begin{figure*}[t]
    \centering
    \subfigure[Example of generated modern dance.]
    {\includegraphics[width=12cm, angle=0]{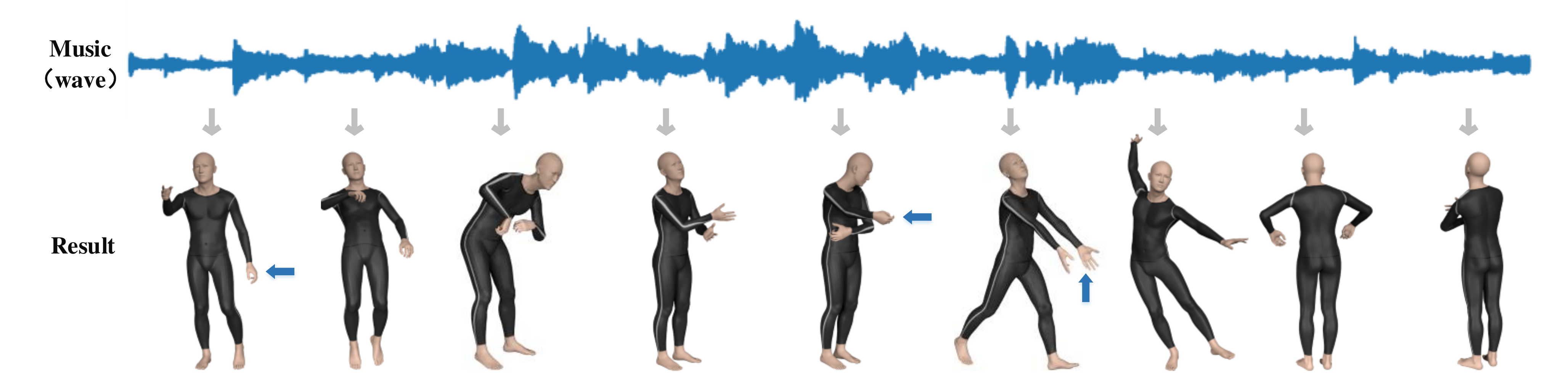}}
    \subfigure[Example of generated curtilage dance.]
    {\includegraphics[width=12cm, angle=0]{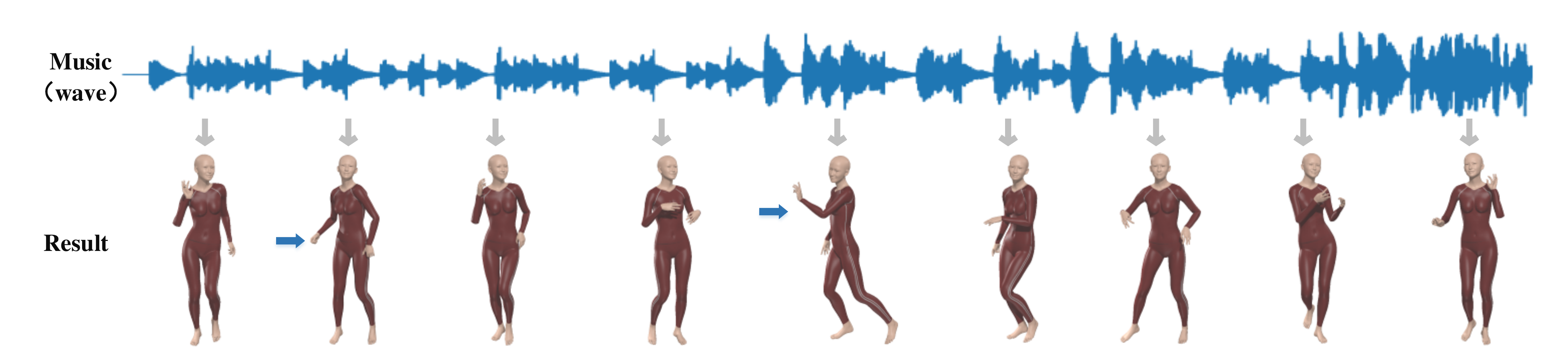}}
    \caption{\textbf{Dance motion generated by our method}. 
    We rendered the generated dance motion with meshes and textures. Our method can obtain 
    realistic, diverse and music-consistent dance motion, and the dance motion includes finger motion (blue arrow). 
    See the video in supplementary materials.}
    \label{fig:qualitative_result}
    \vspace{-2mm}
\end{figure*}

\setlength{\tabcolsep}{4pt}
\begin{table}[t]
\begin{center}
\caption{
\textbf{Comparison of realism}(FID, lower is better), \textbf{diversity}(higher is better), \textbf{rhythm-consistent}(rhythm hit rate, higher is better). 
}
\label{table:comparison_result}
\begin{tabular}{l c c c| c c c}
\toprule
  \multirow{2}{*}{Method} & \multicolumn{3}{c}{Morden Dance} &  \multicolumn{3}{c}{Curtilage Dance}\\ 
   \cmidrule(lr){2-4}\cmidrule(lr){5-7}
   & FID & Diversity &   Rhythm Hit & FID & Diversity &   Rhythm Hit \\
Real Dances & 6.2 & 56.1 & 61.3\% & 5.3 & 48.7 & 70.6\% \\
\midrule
LSTM-autoencoder\cite{tang2018dance} & 82.1 & 18.3 & 11.2\% & 76.4 & 13.8 & 12.9\%\\
Temporal Conv\cite{ginosar2019learning} & 36.7 & 33.4 & 39.8\% & 33.2 & 37.6 & 50.3\%\\
LSTM\cite{lee2018interactive} & 27.8 & 46.5 & 50.8\% & 26.3 & 40.2 & 58.9\%\\
\midrule
Ours(no GMM loss) & 32.3 & 43.1 & 42.8\% & 22.4 & 38.6 & 52.7\%\\
Ours & \textbf{12.5} & \textbf{52.5} & \textbf{58.7\%} & \textbf{8.7} & \textbf{46.9} & \textbf{69.2}\%\\
\bottomrule	
\end{tabular}
\end{center}
\end{table}
\setlength{\tabcolsep}{1.4pt}

\begin{figure}[t] 
  \begin{minipage}[t]{0.45\textwidth} 
  \centering 
    \includegraphics[width=6cm]{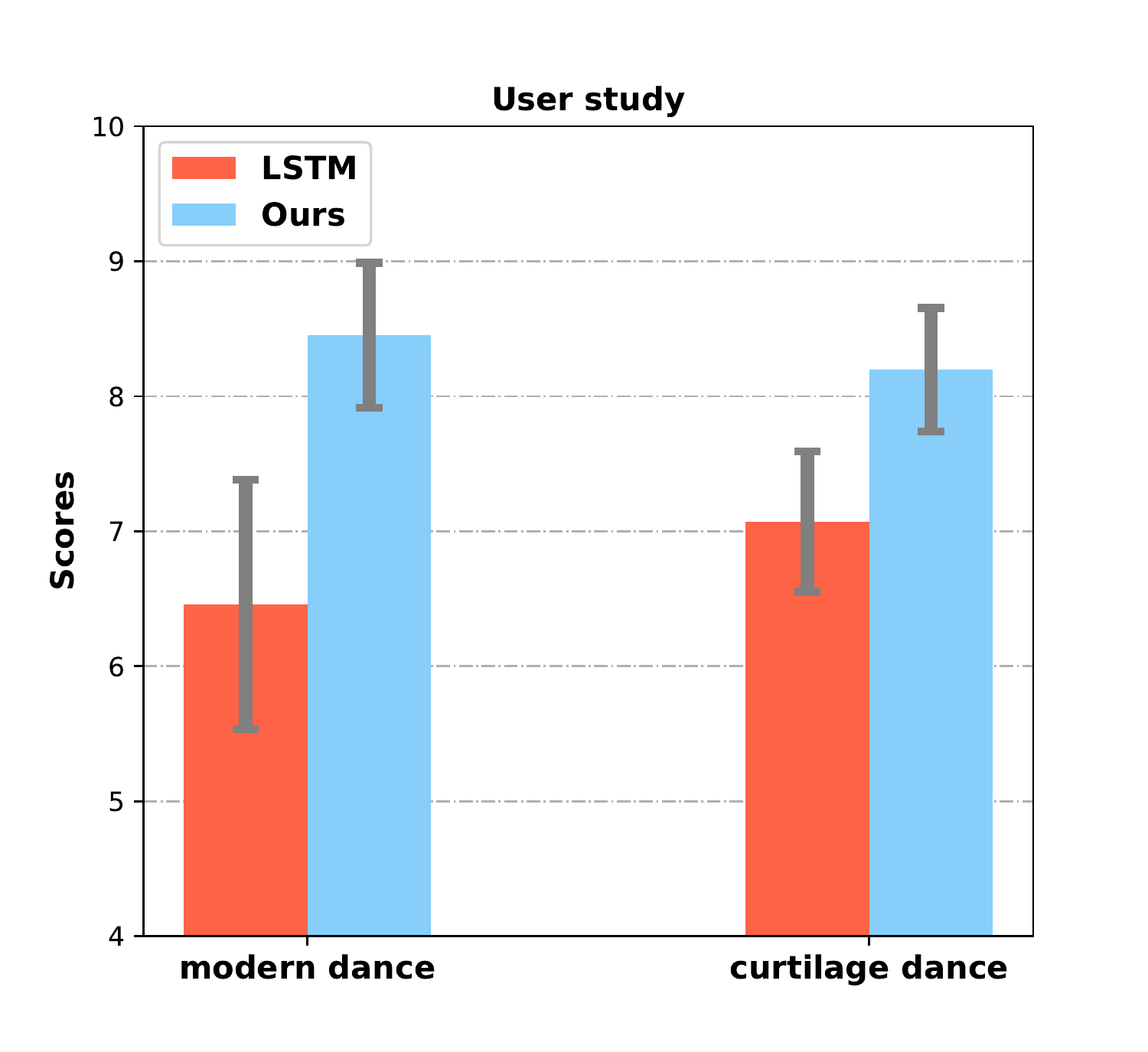} 
    \figcaption{\textbf{Result of user study}. This figure shows the scores given by 25 users, including the means and standrand deviations for dance motion generated by SOTA LSTM\cite{lee2018interactive}(GMM loss) and our method. Our generated dance motions are better than the LSTM.}
    \label{fig:user_study} 
  \end{minipage}%
  \begin{minipage}{0.03\textwidth}
  \
  \end{minipage}
  \begin{minipage}[t]{0.52\textwidth} 
  \centering 
    \includegraphics[width=6.5cm]{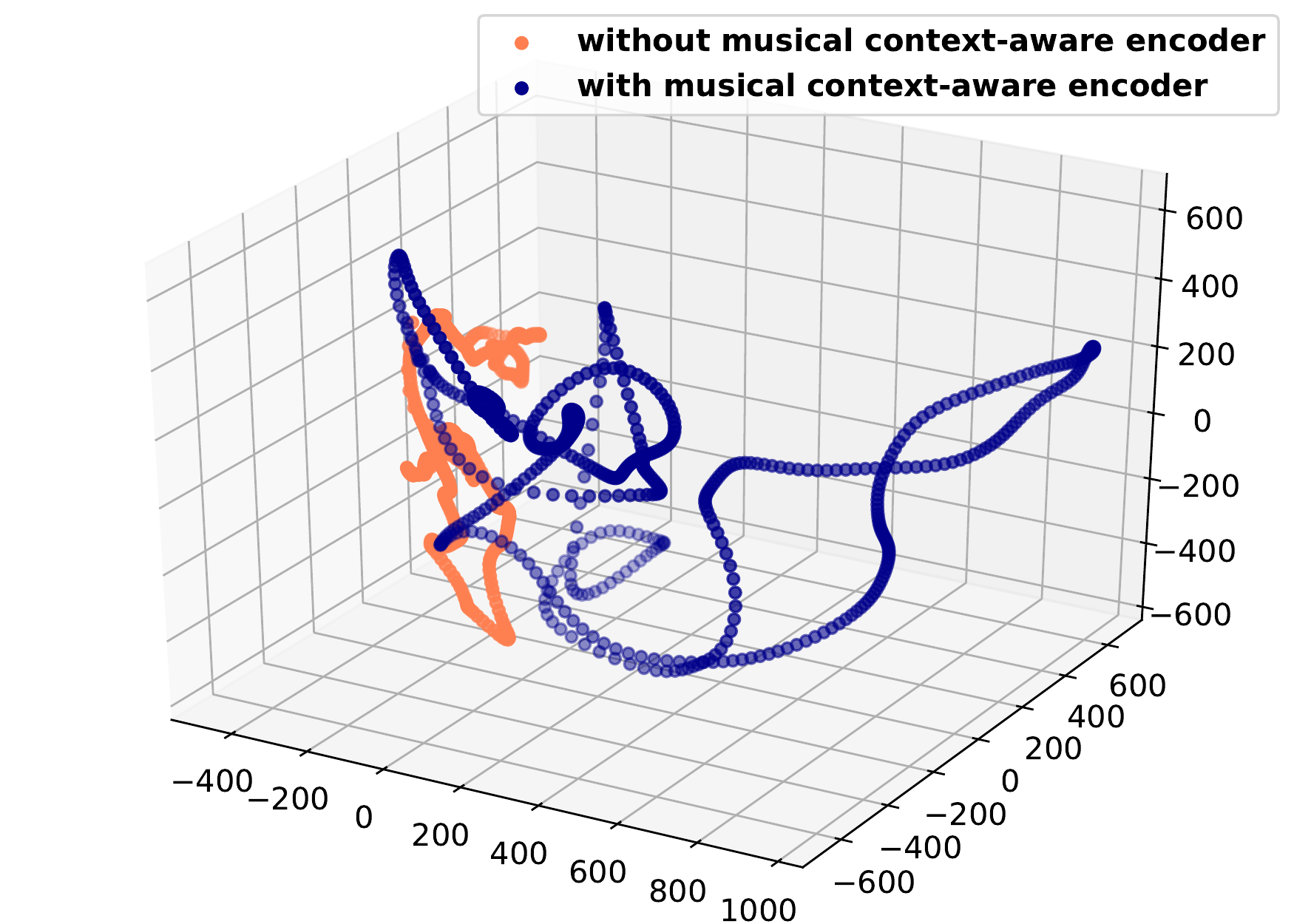} 
    \figcaption{\textbf{The comparison of with/without musical context-aware encoder}. PCA is used to reduce the dimension of end-effector position features, so that we can compare the diversity of generated dance motion. Higher diversity can be obtained when using musical context-aware encoder. }
    \label{fig:pca_musicmap} 
  \end{minipage}%
\end{figure}

\subsection{Result}\label{sec:our_result}
Style-consistency mainly depends on the musical style classification, so we need to make a simple evaluation of our musical style classification model, and then evaluate the performance of the generated dance motion.

\textbf{Musical style classification.}
To verify the effectiveness of our approach, we compared with the Convolution RNN\cite{choi2017convolutional}, and built an FC+Bi-LSTM model as the baseline. The baseline
replaces the temporal convolution layer in our musical style classifier with fully connected layer. The training data 
and strategies are exactly the same as our model.
We use the prediction accuracy to evaluate the final results and our method can obtain excellent results than the convolution RNN and baseline, as shown in Table ~\ref{table:music_attribute}.

\textbf{Evaluation of dance performance.}
We compare against several current SOTA motion modeling methods. 
LSTM-autoencoder\cite{tang2018dance} generates 3D dance motion directly from music. 
Temporal Convolution(Temporal Conv)\cite{ginosar2019learning} models body gestures from speech. LSTM\cite{lee2018interactive} is an autoregressive model, and the generated motion is controlled by control signals. We found that LSTM was difficult to model the dance using MSE loss, so replaced with GMM loss.
20 dance motions(10 modern dance motions and 10 curtilage dance motions) are generated by our model and the above methods, respectively.
We compare them by the realism, diversity, style-consistency, rhythm-consistency, melody-consistency. 

\textbf{1)Realism and style-consistency.}
We evaluate the dance motion realism and style-consistency by Fr\'echet Inception Distance(FID)\cite{heusel2017gans}, similar to \cite{yan2019convolutional}. Because FID requires an action classifier to extract dance features, we train an action classifier based on temporal convolution and Bi-LSTM on our dataset as the feature extractor.
The realism is reflected in that the generated dance needs to be close to the real dance, and the style-consistency is reflected in the classifier is classified according to the dance type. As shown in Table~\ref{table:comparison_result}, the FID of our generated dance motions is lower, which means our results is closer to the real dances and more style-consistent.
\textbf{2)Diversity.} We evaluate the dance motion diversity by 
the average feature distance among different dance motions, similar to \cite{lee2019dancing}. Our method can generate more diverse dance motions, as shown in Table~\ref{table:comparison_result}. 
\textbf{3)Rhythm-consistency.} In Sec~\ref{sec:music_feature_extraction}, we analyze rhythm-consistency, so we use the rhythm hit rate as the evaluation method of rhythm-consistency(Rhythm hit needs to meet the error within 0.25s), similar to \cite{lee2019dancing}. The comparison shows our method can obtain higher rhythm hit rate in Table~\ref{table:comparison_result}.

Our method is superior to other methods by the above quantitative evaluation. Compared against the LSTM-autoencoder\cite{tang2018dance}, generates 3D dance motion directly from music, our method is completely superior to it. More importantly, we can generate different dance types with the same model, but they can not. Our DanceNet is based on temporal convolution, and we combine dilated temporal convolution with the gated activation unit to obtain better performance than general temporal convolution\cite{ginosar2019learning}. We compared against the autoregressive model, LSTM(GMM loss)\cite{lee2018interactive}, and our results are better than theirs. Our explanation is that the autoregressive model based on dilated temporal convolution is better and more robust than LSTM-based for modeling dance  motion.



\textbf{User study.} Because some indicators are difficult to quantify, e.g., melody-consistency, and each of the above evaluation methods is based on one characteristic, it is difficult to fully evaluate result. Therefore, we use user study to comprehensively evaluate the generated dance motions. Evaluating all results would take a lot of time for the users, so we only evaluate our results with LSTM\cite{lee2018interactive}(better than other methods).
We asked 25 users to score these dance motions. The score basis consists of the realism, diversity, music-consistency. More scoring indicators are described in supplementary materials. 
We report the mean scores and standrand deviations for the dance motions generated by
our model and the LSTM, as shown in Figure ~\ref{fig:user_study}. The result of user study shows 
that our generated dance motions are obviously superior to the LSTM.
Our mean score reaches 8.452 (modern dance), 8.196 (curtilage dance), and the standrand deviations
are significantly smaller than the LSTM, especially modern dance.
In our motion data, modern dance is more diverse than curtilage dance, which is a very important reason that the modern dance generated by the LSTM is worse than curtilage dance(lower mean score, larger standard deviation). Our approach can generate diverse modern dance, and the score is slightly higher than curtilage dance. It means that our method is more robust to diverse(complex) dance motion. In addition, we rendered the generated dance motions with meshes and textures. We show two examples in Figure ~\ref{fig:qualitative_result}, one for modern dance and another for curtilage dance. Both examples show that our method can obtain 
realism, diverse and music-consistent dance motion.


\subsection{Discussion}\label{sec:discuss}
To demonstrate our method, we need to discuss every part of our method, including: mel spectrum v.s. music features(onset,beat,chroma), musical context-aware encoder, with/without end-effector position, MSE loss v.s. GMM loss. We perform qualitative evaluation in this section, and more quantitative results are shown in the supplementary materials.


\begin{figure}[t] 
  \begin{minipage}[t]{0.49\textwidth} 
  \centering 
  \subfigure[left hand]
  {\includegraphics[width=2.9cm, angle=0]{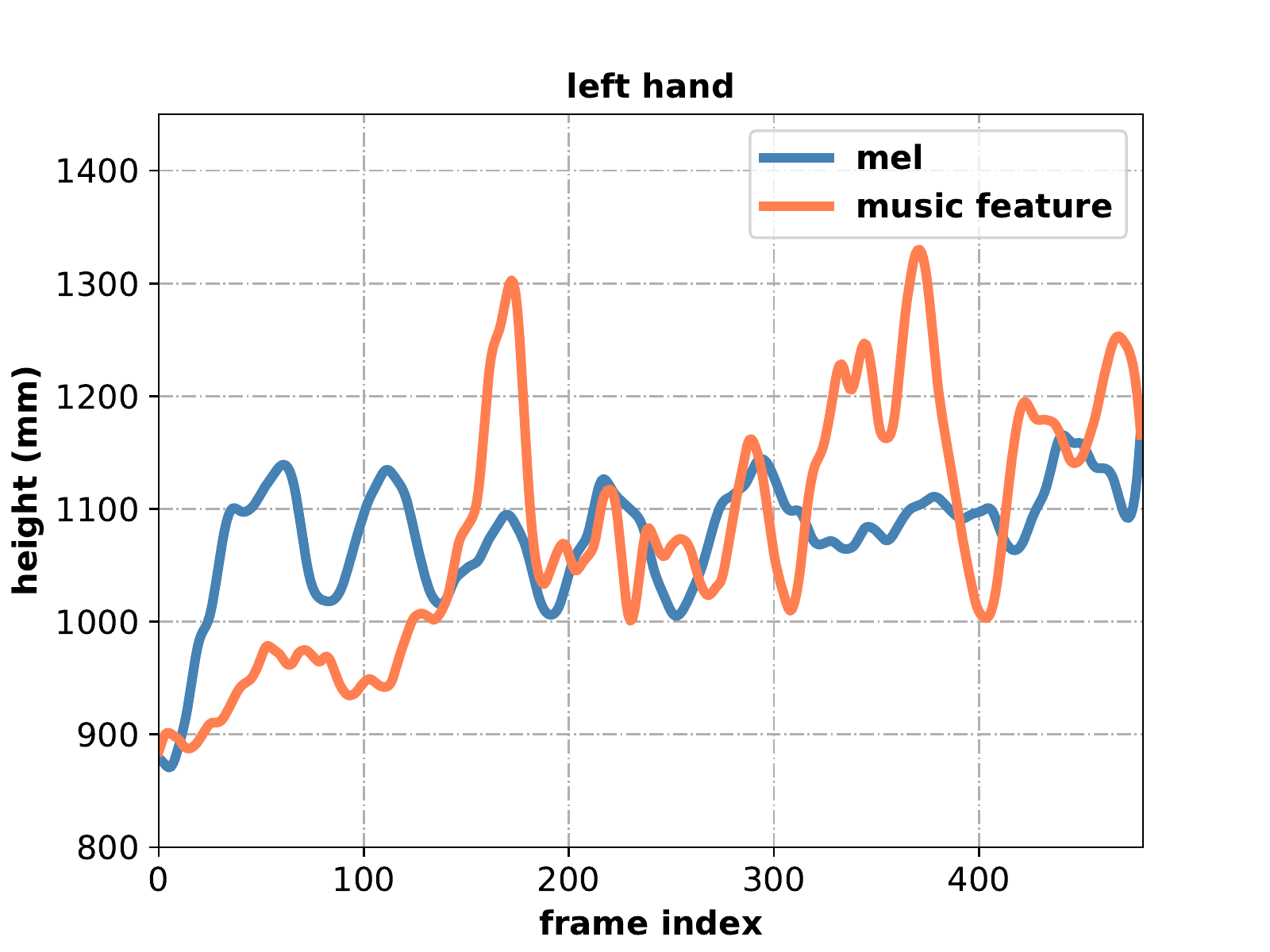}}
  \subfigure[right hand]
    {\includegraphics[width=2.9cm, angle=0]{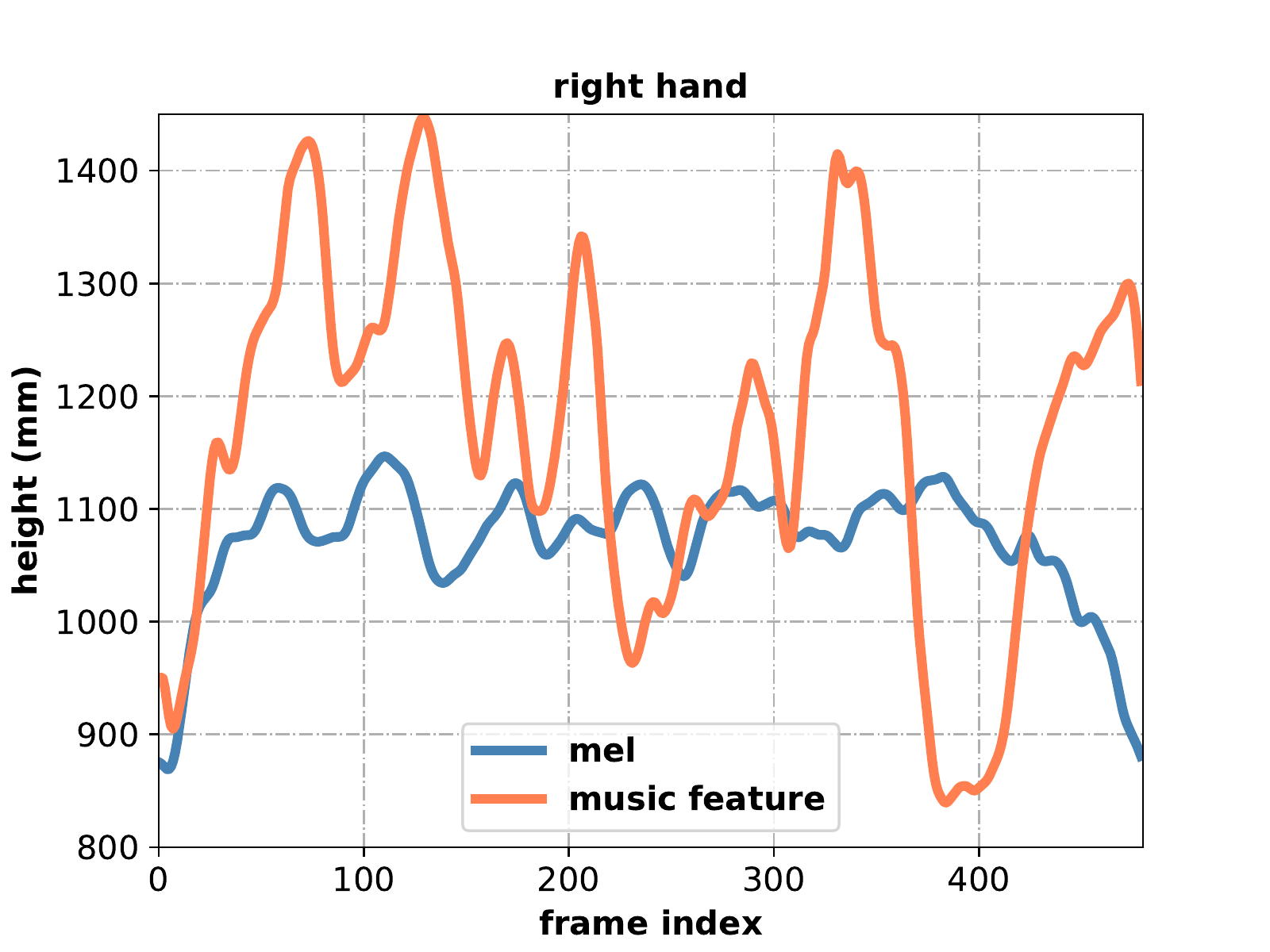}}
  \figcaption{\textbf{The comparison between mel spectrum and music features}. The generated dance has a richer variation of left/right hand motion with music features as input, especially right hand motion.}
 \label{fig:mel_music_hand_height}
  \end{minipage}%
  \begin{minipage}{0.02\textwidth}
  \
  \end{minipage}
\begin{minipage}[t]{0.49\textwidth} 
    \centering
    \subfigure[left toe\_end]
    {\includegraphics[width=2.9cm, angle=0]{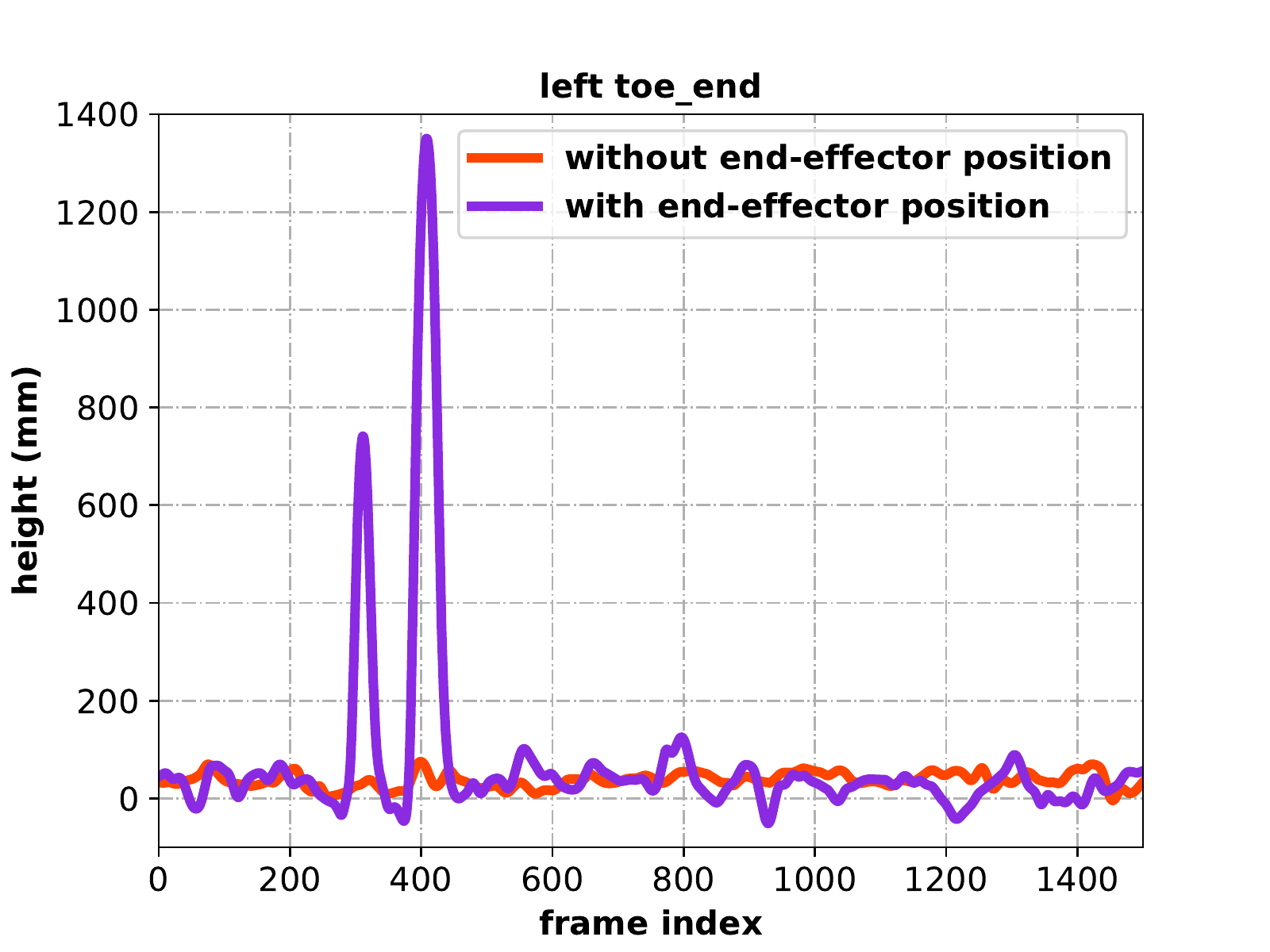}}
    \subfigure[right toe\_end]
    {\includegraphics[width=2.9cm, angle=0]{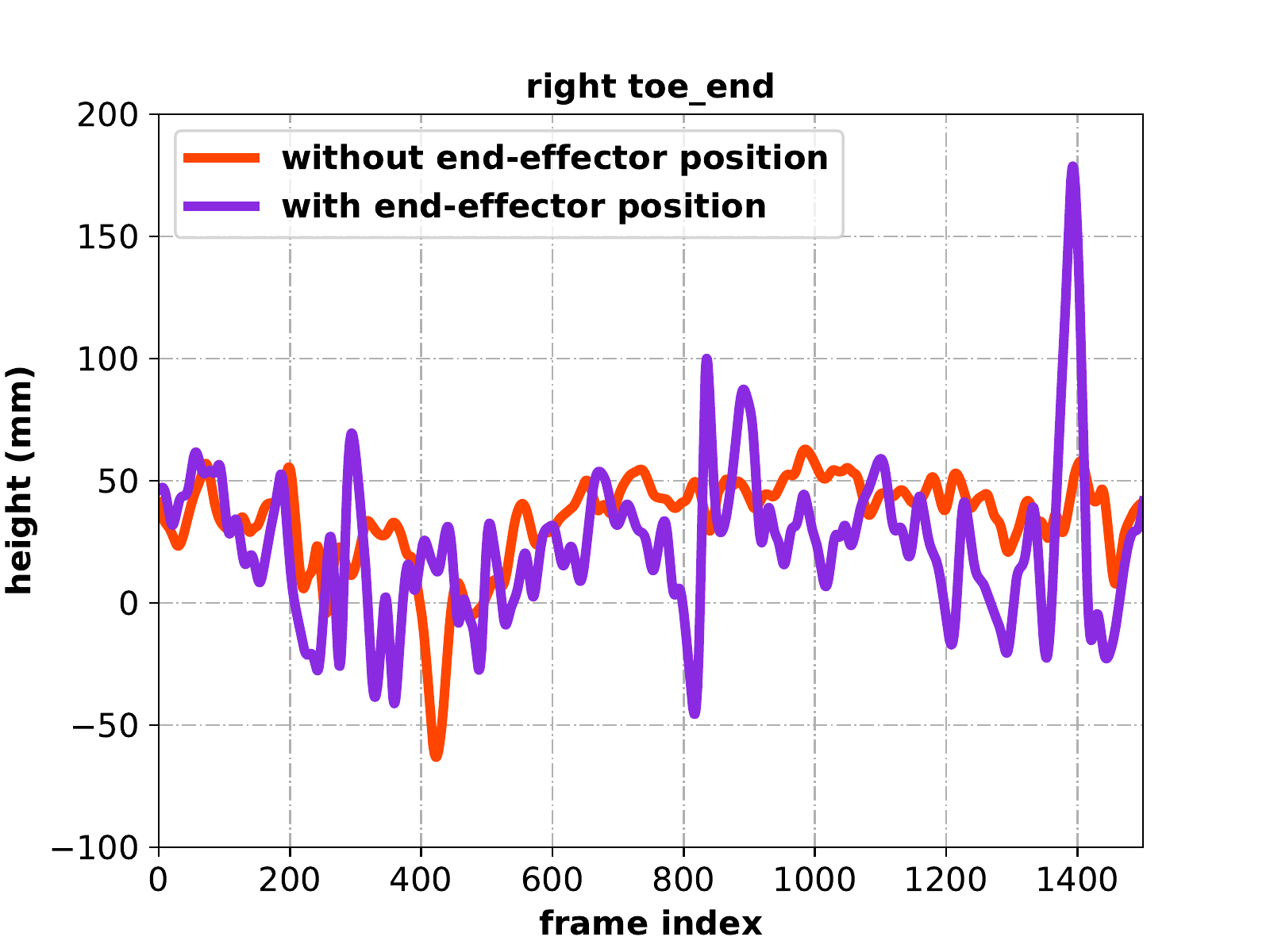}}
    \figcaption{\textbf{The comparison of whether add end-effector position to the 
    motion feature}. Left/right 
    toe\_end motion generated by the model with end-effector position has more variety.
    }
    \label{fig:end_position_foot_height}
  \end{minipage}%
\end{figure}


\textbf{Mel spectrum v.s. Music features.}
We trained the model with mel spectrum and music features respectively, and tested on 
the same music clip. When we use the mel spectrum as input,
the generated dance motion is very stiff and appears jitter problem
(we smooth the motion via Gaussian filter with a large kernel size $\sigma =4$). 
We compared the generated motion generated by the same music clip 
and plotted the height of the left/right hand over time (480 frames), as shown in 
Figure ~\ref{fig:mel_music_hand_height}. When the music features are used as input, 
the generated dance is more diverse and music-consistent.
It means that the music features in our method have higher generalization ability. 


\textbf{Musical context-aware encoder.}
The musical context-aware encoder is a very important part. We tried to directly take the music features as 
control signals without the musical context-aware encoder, 
and adopted the same training data and strategies to train the model. 
We found that the realism and diversity of the dance motion generated by the model is poor (only simple dance steps or standing still).
In order to better compare the results, 
we extracted the end-effector position (important motion feature) of the generated motion and 
used PCA to reduce the dimension
to visualize the results,  as shown in Figure ~\ref{fig:pca_musicmap}.
It is obvious that we can get more realistic and diverse motion when we use musical context-aware encoder. One important reason is that music and dance are two modalities and the music features should be encoded. Another explanation is that the dance motion is smooth, the input control signal should be smooth, and the jitter control signals(music features) reduce the realism. Obviously, we visualized the music-context-code and found that it is smooth.


\textbf{With/Without end-effector position.}
In Section ~\ref{sec:motion_representation}, we explain why the end-effector position feature is added to the motion feature.
 In order to verify its advantages, we trained a model without the end-effector position 
 feature. 
 We compared the dance motion generated by the same music clip and plotted the height of 
 the toe\_end position over time, as shown in Figure ~\ref{fig:end_position_foot_height}.
It shows that adding the end-effector position can get more diverse dance motions. Eliminating the accumulation errors from root to end-effector can predict more accurate dance motion, thereby increasing the diversity.

\textbf{MSE loss v.s. GMM loss.} 
In the existing motion modeling methods\cite{li2017auto}\cite{lee2018interactive}, MSE loss is usually used. In Sec~\ref{sec:our_result}, we mentioned it is difficult to train the autoregressive model LSTM on our dataset using MSE loss. Similarly, we used MSE loss to train our DanceNet and found that the modeling ability was poor(Table~\ref{table:comparison_result}). Dance is a long sequence of motion. The MSE loss would cause the model to predict a certain motion frame, which could cause error accumulation. In addition, dance is more diverse than simple locomotion, and MSE loss reduces the diversity. The GMM loss allows DanceNet to model a probability distribution that can cover more fileds in the motion graph. We can sample around the predicted mean motion($\hat{\mu }$ in Sec~\ref{sec:gmm_loss}) to increase diversity in the generation phase.
\section{Conclusion}\label{sec:conclusion}

In this paper, we propose a novel autoregerssive model, DanceNet, and it can generate realistic, diverse and music-consistent dance motion from the input music. In addition, DanceNet can generate dance motion of different dance types from the same model. To train our model, we build a high-quality music-dance pair dataset, including two dance types. Our results demonstrate the power of our method.
However, there are still some defects: the generated dance motions are not as 
good as professional dancers, and there is slight foot sliding.
These are the focus of our future work.

\clearpage
%
%
\bibliographystyle{splncs04}
\bibliography{egbib}

\newpage

\textbf{\large Music2Dance: DanceNet for Music-driven Dance Generation
(Supplementary)}

\setcounter{section}{0}

\section{Implementation details}\label{sec:implementation}
\textbf{Musical style classification}.
We divided the music data into training data (about 82.5\%, 
musical style 1 \verb'curtilage dance': 14.5 hours, musical style 2 \verb'modern dance': 15 hours) 
and test data (17.5\%, musical style 1: 3 hours, musical style 2: 3.2 hours). 
We used Adam \cite{kingma2014adam} to optimize the model with a batch size of 128 for 30 epochs. The initial learning rate is  $1\times 10^{-4}$ and is dropped by 10 at 18th and the 25th epoch.

\begin{figure}
	\centering
	\includegraphics[width=\linewidth]{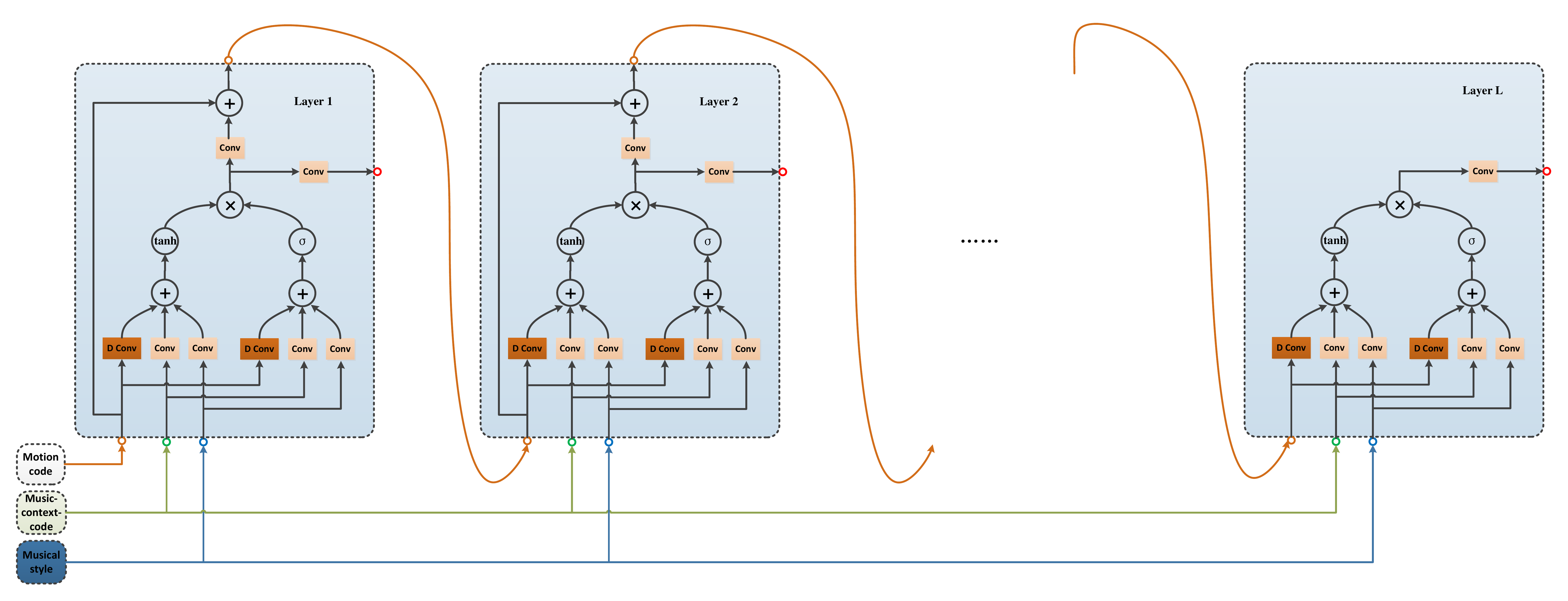}
	\caption{\textbf{Residual motion control stacked module}. We stack L layers to build our module.  The "Conv" is 1D convolution, and the "D Conv" is dilated convolution.}
	\label{fig:key_module}
\end{figure}

\subsection{DanceNet}
\textbf{Residual motion control stacked module}.
In order to better describe our key module: residual motion control stacked module, we elaborate it in more detail, as shown in Figure~\ref{fig:key_module}. We stack L layers to build our module, and L is 20 in our experiment. The dilated coefficients are: $2^0, 2^1, 2^2, 2^3, 2^4, 2^0, ... $. The kernel size of the dilated convolution is set to 2. So the receptive
field in our model is 126 frames. The receptive field is obtained by considering the performance of generated dance and the efficiency in inference phase.

\textbf{Data clustering}. We need to cluster the dance data because the data distribution 
is not uniform. It can help improve the variety of generated dance motion. 
In training phase, the window size of motion sample is 480 frames, and we adopt k-means \cite{macqueen1967some} 
to cluster motion samples. It is mainly noted that the motion feature adopted by k-means clustering 
is not joint rotation, but the joint position feature. After the clustering results are obtained, 
the training samples are sampled by sliding window according to the category probability of each sample. 
The window size is 480 and the stride is 3 frames.

\textbf{Training details}. Gaussian noise is 
added to input and ground-truth, 
and dropout (0.4) is used for input motion feature. We used Xavier normal 
\cite{glorot2010understanding} 
to initialize our model, and RMSprop optimization \cite{tieleman2012lecture}. The DanceNet was trained 1300 epochs, the learning rate is initialized to $4\times 10^{-4}$ and is dropped by 10 at 1000th epoch.

\setlength{\tabcolsep}{4pt}
\begin{table}[t]
\begin{center}
\caption{
\textbf{Ablation Study.}
\textbf{Comparison of realism}(FID, lower is better), \textbf{diversity}(higher is better), \textbf{rhythm-consistent}(rhythm hit rate, higher is better).
}
\label{table:dis_result}
\begin{tabular}{l c c c| c c c}
\toprule
  \multirow{2}{*}{Method} & \multicolumn{3}{c}{Morden Dance} &  \multicolumn{3}{c}{Curtilage Dance}\\ 
   \cmidrule(lr){2-4}\cmidrule(lr){5-7}
   & FID & Variety &   Rhythm Hit & FID & Variety &   Rhythm Hit \\
\midrule
Mel spectrum & 25.8 & 28.3 & 34.4\% & 18.7 & 23.2 & 38.5\%\\
Onset & 15.8 & 46.5 & 60.9 & 13.6 & 39.8 & \textbf{70.8\%}\\
Onset+beat & 14.9 & \textbf{52.9} & \textbf{61.5} & 9.2  & 43.2 & 69.7\%\\
Onset+beat+chroma & \textbf{12.5} & 52.5 & 58.7\% & \textbf{8.7} & \textbf{46.9} & 69.2\%\\
\midrule
\midrule
\tabincell{l}{w/o musical \\ context-aware encoder}  & 22.1 & 50.4 & 55.6\% & 15.4 & 42.6 & 64.5\%\\
\midrule
\tabincell{l}{w/o end-effector \\ position}  & 13.6 & 43.2 & 53.7\% & 9.7 & 36.4 & 65.8\%\\
\midrule
Ours & \textbf{12.5} & \textbf{52.5} & \textbf{58.7\%} & \textbf{8.7} & \textbf{46.9} & \textbf{69.2}\%\\
\bottomrule	
\end{tabular}
\end{center}
\end{table}
\setlength{\tabcolsep}{1.4pt}

\section{Discussion details}\label{sec:discussion_details}
\textbf{Music features.} In addition to comparing with the mel spectrum, we perform an ablation study on our music features, as shown in Table~\ref{table:dis_result}. Our method is significantly better than the Mel spectrum. Comprehensively considering the realism, diversity and rhythm-consistency, we adopt onset, beat and chroma as the music features to represent the music. 

\textbf{Musical context-aware encoder.} The result(without the encoder) is shown in Table~\ref{table:dis_result}, indicating that the dance motion is unrealistic and the diversity is poor. The reason is that the music and dance are two modalities, and it is difficult to fuse features without the musical context-aware encoder.

\textbf{End-effector position.} If the end-effector position is not added to the motion features, the diversity of dance motion predicted by the model is limited (shown in Table~\ref{table:dis_result}). The explanation is that the predicted motion is not accurate due to the error accumulation from root to end-effector, resulting in only repeating some simple dance steps.

\setlength{\tabcolsep}{4pt}
\begin{table}[t]
\begin{center}
\caption{
Post-processing comparision on testing data (MSE).
}
\label{table:post_processing}
\begin{tabular}{l|c}
\hline
Method & Mean Squared Error (MSE)  \\
\hline
Baseline (LSTM) & 0.0028   \\
\textbf{Ours (Temproal Conv)} & \textbf{0.0010} \\
\hline

\end{tabular}
\end{center}
\end{table}
\setlength{\tabcolsep}{1.4pt}

\section{Post-processing}\label{sec:post_process_supp}
Foot sliding is a common problem in motion generation.
Similar as other methods \cite{min2012motion}\cite{xia2015realtime}\cite{holden2016deep}\cite{lee2018interactive}, 
we first attempt to solve this problem with IK, 
which requires very high accuracy of the predicted foot constraints $\hat{c}_{foot}$ ($> $ 95\%).
However, we find that the accuracy of the predicted foot constraints $\hat{c}_{foot}$ is not high 
sufficiently (about 85\%-90\%) due to the diversity and complexity of dance motion. 
The jitter problem occurs if we adopt IK. Therefore, we propose 
a Foot Constraint Model to reduce the problem.
To our knowledge, this is the first time using the network to solve the foot sliding.

\textbf{Foot Constraint Model.}
Inspired by temporal convolution\cite{pavllo20193d}, our Foot Constraint Model consists of 3 temporal convolution layers, 
and the kernel size is set to 3 with dilation 1, 3 and 9, respectively. 
The goal of the Foot Constraint Model is to solve the foot sliding of the motion $\hat{x}$ 
generated by the DanceNet, so we only deal with the motion of the lower body (including root joint). 
The input of the model consists of 
the joint rotation motion of the lower body $x_{lrot}^{'}$, the position of foot end-effectors
(left/right $toe\_end$) $p_{fend}^{'}$, 
and the foot constraints ${c}_{foot}^{'}$. The output is the increment of the lower body joint rotation, 
so it can be described as:
\begin{equation}
	\hat{x}_{lrot}^{'}=x_{lrot}^{'}+F_{post}(x_{lrot}^{'},p_{fend}^{'},{c}_{foot}^{'})
	\label{equ:post_model}
\end{equation}

$F_{post}$ is the Foot Constraint Model. In the training phase, 
we simulate the foot sliding data  by adding gaussian noise 
and gaussian smoothing into the ground-truth data. 
Finally, we adopt MSE loss function, 
and we add the smoothing loss (smoothing factor is set to 0.1 in our experiment).

\textbf{Training details.}
We directly simulated the dance motion to get the foot sliding motion.
Our dataset was divided into training data (85\%) and test data (15\%).
Similarly, the training sample is sampled by sliding window with size 480 frames, stride 3. 
The model is optimized by Adam \cite{kingma2014adam} with 200 epochs, 
and initial learning rate is $5\times 10^{-4}$.

\textbf{Result.}
Our Foot Constraint Model is a stacked temporal convolution network (3 layers). 
To illustrate its effectiveness, we propose a baseline: 
1 fully connected layer + LSTM + 1 fully connected layer. 
We compare the performance on the test data, 
and use MSE for evaluation, and our method can obtain better performance, as shown in 
Table ~\ref{table:post_processing}. 

The motion representation in the LSTM-based method\cite{lee2018interactive}(trained by our GMM loss) is consistent with ours, so the results of our DanceNet and the LSTM are post-processed using the Foot Constraint Model, respectively.

\setlength{\tabcolsep}{4pt}
\begin{table}[t]
\begin{center}
\caption{
Scoring level. H:high, M:middle, L:low, N:not
}
\label{table:score_level}
\begin{tabular}{c|c}
\hline
Score level & Content description  \\
\hline
0 & completely do not move with music   \\
2 & realistic(N)  \\
4 & realistic(L), music-consistent(N)  \\
6 & realistic(L), music-consistent(N), diverse(N), foot sliding(H) \\
8 & realistic(M), music-consistent(L), diverse(L), foot sliding(H)  \\
9 & realistic(H), music-consistent(M), diverse(M), foot sliding(L) \\
10 & realistic(H), music-consistent(H), diverse(H), foot sliding(N)  \\
\hline

\end{tabular}
\end{center}
\end{table}
\setlength{\tabcolsep}{1.4pt}
\section{User study Scoring}\label{sec:user_study_score}
We asked 25 people to score these dance motions. Score basis: 
the realism (25\%), diversity (25\%), music-consistency (40\%), 
foot sliding (10\%). More scoring indicators are described in Table ~\ref{table:score_level} 
\end{document}